\documentclass[10pt,twocolumn,letterpaper]{article}

\usepackage[pagenumbers]{cvpr} 

\usepackage[dvipsnames]{xcolor}
\usepackage{calc}

\definecolor{tabfirst}{rgb}{1, 0.7, 0.7}
\definecolor{tabsecond}{rgb}{1, 0.85, 0.7}
\definecolor{tabthird}{rgb}{1, 1, 0.7}
\definecolor{tabgray}{rgb}{0.9, 0.9, 0.9}
\definecolor{turquoise}{cmyk}{0.65,0,0.1,0.3}
\definecolor{purple}{rgb}{0.65,0,0.65}
\definecolor{darkgreen}{rgb}{0, 0.5, 0}
\definecolor{orange}{rgb}{0.8, 0.6, 0.2}
\definecolor{red}{rgb}{0.8, 0.2, 0.2}
\definecolor{darkred}{rgb}{0.6, 0.1, 0.05}
\definecolor{blueish}{rgb}{0.0, 0.3, .6}
\definecolor{light_gray}{rgb}{0.7, 0.7, .7}
\definecolor{pink}{rgb}{1, 0, 1}
\definecolor{greyblue}{rgb}{0.25, 0.25, 1}

\newcommand{\figref}[1]{Fig.~\ref{fig:#1}}
\newcommand{\tabref}[1]{Table~\ref{tab:#1}}

\newcommand{\secref}[1]{Sec.~\ref{sec:#1}}

\def\eg{\emph{e.g}\onedot}

\def\ie{\emph{i.e}\onedot}

\newcommand\lft{\mathopen{}\left}
\newcommand\rgt{\aftergroup\mathclose\aftergroup{\aftergroup}\right}

\newcommand{\myparagraph}[1]{ \vspace{3pt}  \noindent {\bf #1}\,\,\,}

\newcommand{\cmark}{\ding{52}}
\newcommand{\greencmark}{\color{darkgreen}{\cmark}}
\newcommand{\xmark}{\ding{56}}
\newcommand{\redxmark}{\color{darkred}{\xmark}}

\newcommand{\ncond}{M}
\newcommand{\ntgt}{N}

\newcommand{\Image}{\mathrm{I}}
\newcommand{\Time}{t}
\newcommand{\TimeSet}{\mathrm{T}}
\newcommand{\Camera}{\mathrm{P}}
\newcommand{\cond}{\text{cond}}
\newcommand{\tgt}{\text{tgt}}

\newcommand{\condImage}{\Image^\cond}
\newcommand{\condCamera}{\Camera^{\cond}}
\newcommand{\condTime}{\TimeSet^{\cond}}

\newcommand{\tgtImage}{\Image^\tgt}
\newcommand{\tgtCamera}{\Camera^\tgt}
\newcommand{\tgtTime}{\TimeSet^\tgt}

\newcommand{\unet}{\epsilon_\theta}
\newcommand{\latentT}{z^{\tgt}(i)}
\newcommand{\cfgCondTime}{\mathrm{c}_\TimeSet}
\newcommand{\cfgCondImagePose}{\mathrm{c}_\Image}
\newcommand{\cfgScaleTime}{s_\TimeSet}
\newcommand{\cfgScaleImagePose}{s_\Image}

\newcommand{\gridImage}{\mathrm{G}}

\definecolor{cvprblue}{rgb}{0.21,0.49,0.74}
\usepackage[pagebackref,breaklinks,colorlinks,allcolors=cvprblue]{hyperref}
\usepackage{multirow}
\usepackage{tabu}
\usepackage{xcolor}
\usepackage{colortbl}
\usepackage{makecell}
\usepackage{booktabs}
\usepackage{pifont}
\usepackage{bbm}
\usepackage{graphicx}
\usepackage{algorithm}
\usepackage{algpseudocode}

\def\paperID{****} 
\def\confName{CVPR}
\def\confYear{2025}

\title{CAT4D: Create Anything in 4D with Multi-View Video Diffusion Models}

\author{
Rundi Wu$^{1,2}$
\quad
Ruiqi Gao$^1$
\quad
Ben Poole$^1$
\quad
Alex Trevithick$^{1,3}$\\
\quad
Changxi Zheng$^2$
\quad
Jonathan T. Barron$^1$
\quad
Aleksander Hołyński$^{1}$
\vspace{2mm}
\\
\centerline{$^1$Google DeepMind \quad $^2$Columbia University\quad $^3$UC San Diego}
}

\begin{document}

\twocolumn[{%
\renewcommand\twocolumn[1][]{#1}%
\maketitle
\begin{center}
\vspace{-0.3in}
    \includegraphics[width=0.99\linewidth]{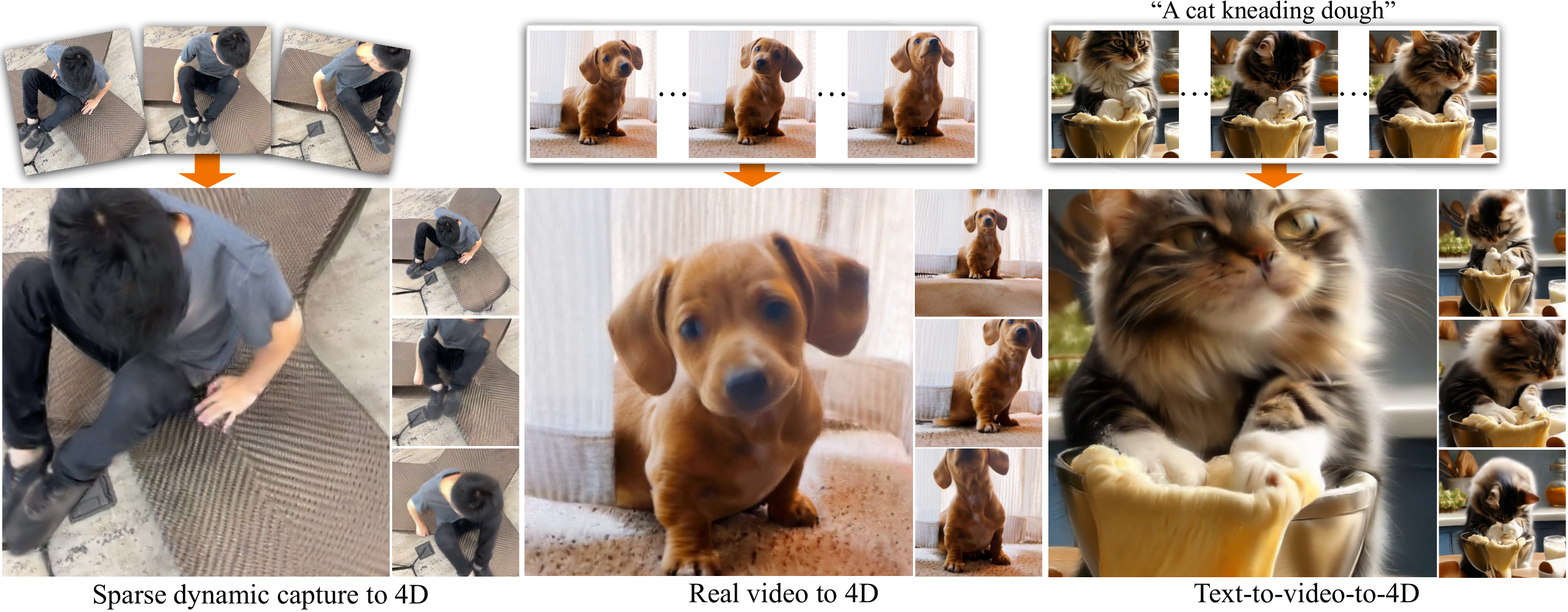} 
    \vspace{-0.05in}
    \captionof{figure}{CAT4D enables 4D scene creation from any number of real or generated images or video frames.} 
    \label{fig:realteaser}
\end{center}
}]

\begin{abstract}
We present CAT4D, a method for creating 4D (dynamic 3D) scenes from monocular video.
CAT4D leverages a multi-view video diffusion model trained on a diverse combination of datasets to enable novel view synthesis at any specified camera poses and timestamps. Combined with a novel sampling approach, this model can transform a
single monocular video into a multi-view video, enabling robust 4D reconstruction
via optimization of a deformable 3D Gaussian representation.  We demonstrate competitive performance on novel view synthesis and dynamic scene
reconstruction benchmarks, and highlight the creative capabilities for 4D scene generation from real or generated videos.
See our project page for results and interactive demos: \url{cat-4d.github.io}.

\end{abstract}    
\section{Introduction}
\label{sec:intro}

\begin{figure*}
    \centering
    \includegraphics[width=0.99\linewidth]{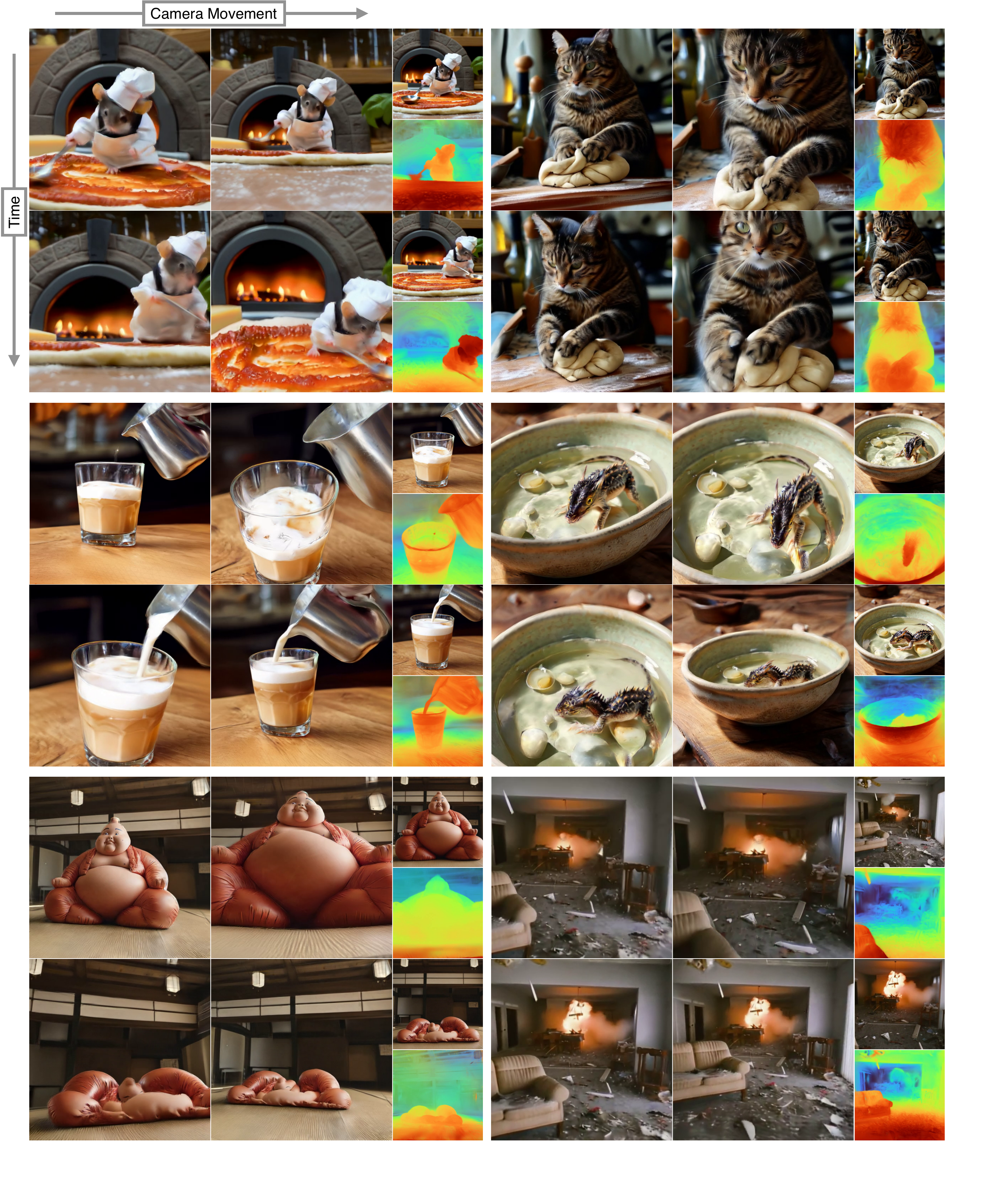} 
    \caption{\textbf{Qualitative results:} CAT4D can generate high-quality dynamic 3D scenes from a single input monocular video. For each example, we show four rendered images, varying in time along the vertical axis, and varying in viewpoint along the horizontal axis. We also show a depth map (bottom right) and a frame from the input video (top right) at the same timestamp as the second column of renders.}
    \label{fig:4dgen_gallery}
\end{figure*}

While the world is a dynamic 3D environment, the images and videos we capture represent only a partial snapshot of this reality. Transforming this limited information into an accurate model of the dynamically changing 3D world remains an open research challenge, and progress in this space could enable applications in robotics, film-making, video games, and augmented reality. 

Advances in 3D reconstruction have enabled the creation of accurate 3D and 4D models when certain careful capture requirements are followed~\cite{mildenhall2020nerf,kerbl3Dgaussians,luiten2024dynamic,wu20244d,fridovich2023k,cao2023hexplane,pumarola2021d,bemana2020x,li2022neural}. For static 3D reconstruction, this requires having a large set of multi-view consistent images of the 3D scene. This can be achieved through careful capture and restricted settings, which is challenging to accomplish in most environments, but often feasible with enough time and practice. Reconstructing 4D content, however, requires synchronized multi-view video, which is nearly impossible for a typical user. 

To reduce the dependence on these challenging capture requirements, several methods have turned to data-driven approaches leveraging learned 3D generative priors to improve the quality of 3D reconstruction~\cite{gao2024cat3d,sargent2023zeronvs,wu2024reconfusion,hong2023lrm}. Core to the success of these methods is the availability of high-quality multi-view image datasets that enable learning generative priors.
While this data exists in large quantities for static environments, the same cannot be said for 4D, where the complexity of building a synchronized multi-view capture setup constrains both the amount and diversity of available data, and therefore the ability to build a data-driven 4D prior.
Instead, methods have attempted to regularize 4D reconstruction via geometric constraints~\cite{park2021nerfies,park2021hypernerf}, estimated depth maps~\cite{yoon2020novel,gao2021dynamic}, optical flow fields~\cite{li2023dynibar,li2021neural}, or long-term 2D tracks~\cite{wang2024shape,lei2024mosca}.
But even with all these additional signals, the most effective methods still show noticeable artifacts when viewed from novel viewpoints, especially when inputs only partially observe the target scene.

\begin{figure}
    \centering
    \includegraphics[width=0.99\linewidth]{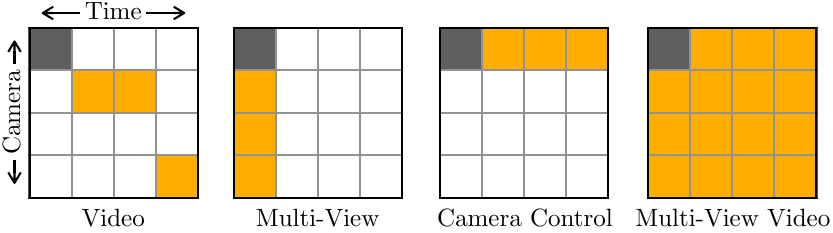}
    \vspace{-0.1in}
    \caption{\textbf{What is a multi-view video model?} Given one or several input images (grey), different generative models have the ability to create novel images (orange) at certain collections of camera viewpoints and timestamps. {\em Video models} generate frames at all timestamps, but without control over camera. {\em Multi-view models} generate at controllable cameras but at a fixed timestamp. {\em Camera-controlled video models} enable the choice of camera per timestamp, but cannot generate multiple cameras per timestamp. {\em Multi-view video models} can generate all views at all timestamps.}
    \vspace{-0.2in}
    \label{fig:camtime}
\end{figure}

In this work, we leverage a variety of real and synthetic data sources to learn a generative prior that enables 4D reconstruction. Inspired by the success of generative priors for sparse-view 3D reconstruction of static scenes~\cite{gao2024cat3d,wu2024reconfusion,sargent2023zeronvs}, we train a multi-view video diffusion model that can transform an input monocular video into \emph{multi-view videos}, which we then use to reconstruct the dynamic 3D scene. Specifically, our model accepts any number of input images captured at different viewpoints and times, and synthesizes as output the scene's appearance at any specified novel viewpoints and novel times (see~\figref{camtime}). Because very little real-world multi-view training data of dynamic scenes exists, we train our model with a bespoke mixture of multi-view images of static scenes~\cite{zhou2018stereo,reizenstein2021common,yu2023mvimgnet,li2019learning}, fixed-viewpoint videos containing dynamics, synthetic 4D data~\cite{greff2022kubric,deitke2023objaverse}, and additional data-sources augmented by a pre-trained video model~\cite{bar2024lumiere} and a multi-view image synthesis model~\cite{gao2024cat3d}.
Our model is trained to produce a consistent collection of frames at specified viewpoints and timestamps, but due resource constraints, it was only trained to generate a finite number of frames at once (16). Since high-quality 4D reconstruction requires many frames at many novel viewpoints, we additionally propose a sampling strategy to generate an unbounded collection of consistent multi-view videos beyond the diffusion model's native output length. 
Finally, we use these generated multi-view videos to reconstruct a dynamic 3D model by optimizing a deformable 3D Gaussian representation~\cite{wu20244d} with a photometric reconstruction loss.

We evaluate our system, CAT4D, on a variety of tasks: 
We first showcase our model's ability to generate images at novel viewpoints and times with disentangled camera and time control, given sparse-view input images at different timestamps.
We then demonstrate our model's ability to perform sparse-view static 3D reconstruction in the presence of scene motion, which is a common failure mode for prior sparse-view reconstruction methods.
We then evaluate on the task of 4D reconstruction from monocular videos with moving cameras, and show that our results are comparable to those from existing state-of-the-art models that critically depend on multiple priors and external sources of information --- all of which our model does not need or use.
Lastly, we show that our model is able to produce compelling 4D generation results from fixed-viewpoint videos that are \emph{scene-scale} (\eg, multiple dynamic objects within dynamic environments, see \figref{4dgen_gallery}), whereas most previous works focus on generating singular objects.

\section{Related Work}
\label{sec:related}

\myparagraph{Dynamic 3D Reconstruction}
Reconstruction and novel view synthesis of dynamic 3D scenes is a challenging research problem.
Many existing methods require multiple synchronized videos captured from different viewpoints as input~\cite{luiten2024dynamic,wu20244d,fridovich2023k,cao2023hexplane,li2022neural,attal2023hyperreel,wang2022fourier,park2021nerfies,park2021hypernerf,kratimenos2025dynmf}, while some more recent works address the more practical setting of casually captured monocular videos as input~\cite{wang2024shape,lei2024mosca,lee2025fast,zhou2024dynpoint,li2023dynibar,li2021neural,zhao2024pseudo}.
These works tend to use radiance field models such as NeRF~\cite{mildenhall2020nerf} or 3DGS~\cite{kerbl3Dgaussians} as the underlying static 3D representations, alongside a deformation field that models the scene motion.
Most recently, Shape-of-Motion~\cite{wang2024shape} and MoSca~\cite{lei2024mosca} greatly improved reconstruction quality by utilizing extra supervision signals estimated from large vision models, such as segmentation masks~\cite{kirillov2023segment,yang2023track}, depth maps~\cite{yang2024depth,piccinelli2024unidepth} and long-term 2D tracks~\cite{doersch2023tapir,karaev2023cotracker}.
However, these methods are not fully automatic (\eg, requiring user clicks for the object mask) and cannot reconstruct regions that are never observed in any input frames.
In contrast, we use a diffusion model to transform an input monocular video to multi-view videos, and directly optimize a deformable 3D Gaussians representation~\cite{wu20244d} without the need of additional supervision signals.

\myparagraph{4D Generation}
Early work for 4D generation used Score Distillation Sampling (SDS)~\cite{poole2022dreamfusion} to distill a 4D representation from image and video diffusion models~\cite{singer2023text}.
Follow up work~\cite{bahmani20244d,ling2024align,zheng2024unified,yu20244real,ren2023dreamgaussian4d,zhao2023animate124,jiang2023consistent4d,zeng2025stag4d,yin20234dgen,bahmani2025tc4d} improved generation quality by performing SDS with a combination of video diffusion models~\cite{singer2022make,blattmann2023stable,menapace2024snap} and multi-view image diffusion models~\cite{liu2023zero,shi2023mvdream,liu2023syncdreamer}.
Recent work sidestepped costly SDS optimization by crafting a sampling scheme of diffusion models that generate multi-view videos~\cite{pan2024fast,yang2024diffusion,li2024vivid}, fine-tuning a video model for 360 spins~\cite{liang2024diffusion4d} or training a feed-forward 4D reconstruction model~\cite{ren2024l4gm}.
However, most prior work focuses on generating a single dynamic object.
Concurrent work DimensionX~\cite{sun2024dimensionxcreate3d4d} achieved 4D scene generation by leveraging multiple LoRAs~\cite{hu2021lora} trained on a video model, each tailored to a specific type of camera motion, to generate multi-view videos, which were then used to reconstruct a 4D scene. However, it is constrained to a fixed set of predefined camera trajectories and only limited results are shown.
In contrast, we train a single multi-view video diffusion model that transforms a monocular video
into multi-view videos at controllable novel viewpoints.

\myparagraph{Video Generation Models with Camera Control}
Our work also relates to video generation models with controllable camera trajectories~\cite{hou2024training,yu2024viewcrafter,guo2023animatediff,sun2024dimensionxcreate3d4d,zhang2024recapture,he2024cameractrl,yang2024direct,kuang2024collaborative,van2024generative,bahmani2024vd3d,zhao2024genxd}. However, these models cannot be used to generate multi-view videos that are consistent with each other.
4DiM\cite{watson2024controlling} trained a diffusion model for synthesizing images under novel views and timestamps, but it's unclear whether their model can produce consistent multi-view videos.
Our system is not intended to be a video generation model, but is instead a dynamic 3D reconstruction and generation system built around a multi-view video diffusion model.
As such, our system has several capabilities that video models lack: \emph{exact} control over camera location and time by rendering the 4D model, real-time rendering, and enabling various visual effects such as object insertion, 3D editing, etc.


\section{Method}
\begin{figure}
    \centering
    \includegraphics[width=0.99\linewidth]{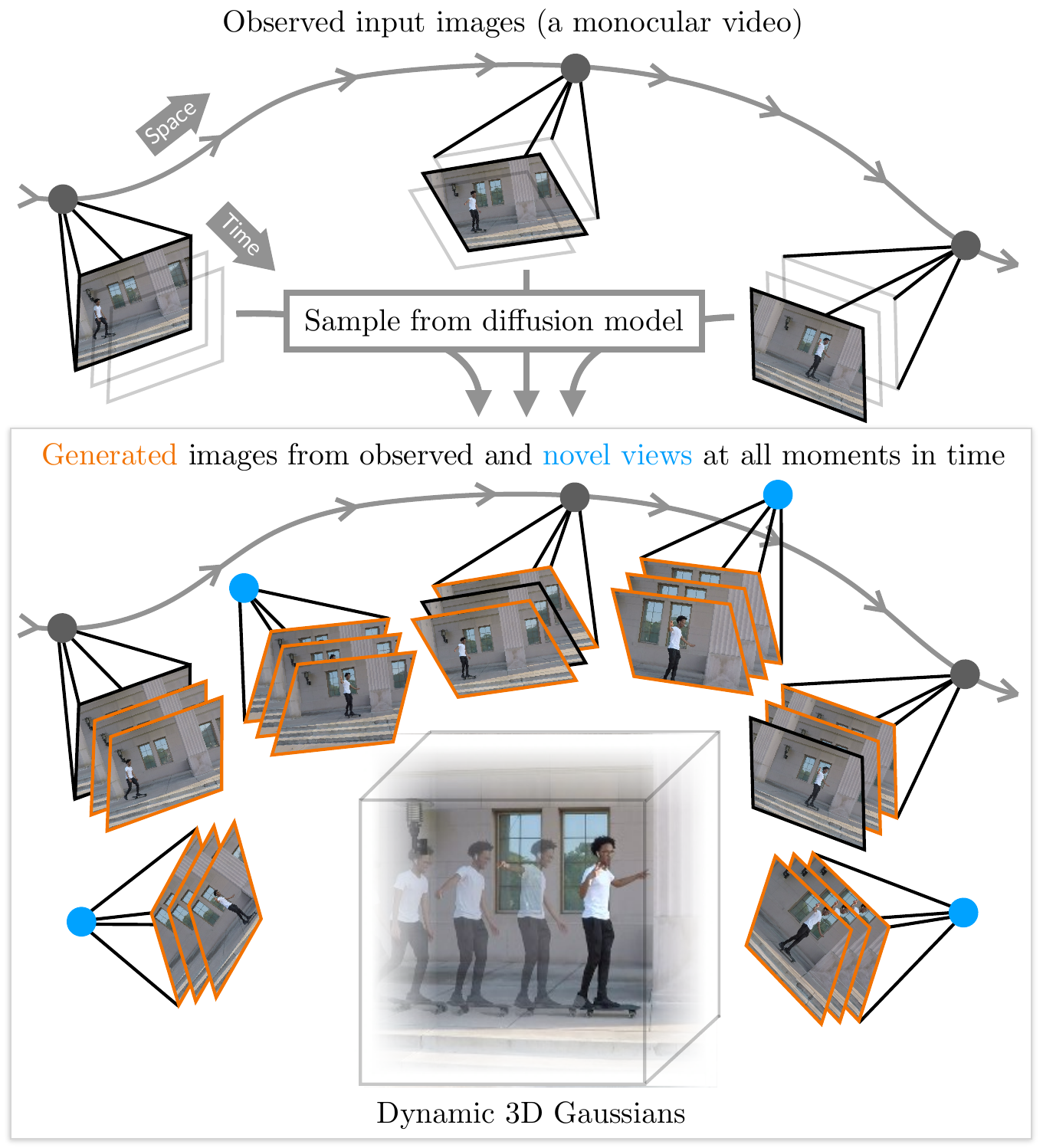}
    \vspace{-0.1in}
    \caption{\textbf{Illustration of the method:} Given a monocular video (top), we generate the missing frames (orange frames) of virtual stationary video cameras positioned at all input poses (gray circles) and novel poses (blue circles) using our multi-view video diffusion model. These frames are then used to reconstruct the dynamic 3D scene as deforming 3D Gaussians. Note that although the input trajectory is visualized with changing viewpoints, our method also works for fixed-viewpoint videos.}
    \vspace{-0.1in}
    \label{fig:overview}
\end{figure}

CAT4D is a two-stage approach for creating a dynamic 3D scene from a monocular input video (see \figref{overview}).
First, we transform the monocular video into a multi-view video using a multi-view video diffusion model.
Then, we reconstruct a dynamic 3D scene by optimizing a deformable 3D Gaussian representation from these generated multi-view videos.
We first describe our multi-view video diffusion model (\secref{method_model}) and our carefully curated training datasets (\secref{method_data}).
We then describe how our model can be used to perform few-view ``bullet-time'' 3D reconstruction (\secref{method_bullettime}).
Lastly, we describe our sampling strategy for generating nearly-consistent multi-view videos (\secref{method_sampling}), and our reconstruction method that yields exactly-consistent dynamic Gaussian models (\secref{method_4d}).

\subsection{Multi-view Video Diffusion Model}
\label{sec:method_model}
We train a diffusion model that takes in a set of views of a dynamic 3D scene (where a ``view'' is an image and its corresponding camera parameters and time) and generates target frames at specified viewpoints and times.
Given $\ncond$ input conditional views with a corresponding set of images $\condImage$, camera parameters $\condCamera$ and times $\condTime$, the model is trained to learn the joint distribution of $\ntgt$ target images $\tgtImage$ given their camera parameters $\tgtCamera$ and times $\tgtTime$,
\begin{equation}
    p\lft(\tgtImage \mid \condImage, \condCamera, \condTime, \tgtCamera, \tgtTime \rgt) \,.
\end{equation}
This model is built on top of CAT3D's diffusion model~\cite{gao2024cat3d}, which is a multi-view latent diffusion model that applies 3D self-attention to connect all image latents.
We adopt an identical architecture while injecting our additional time conditioning into the model.
Specifically, we encode each time condition $\Time \in \condTime \cup \tgtTime$ via a sinusoidal positional embedding~\cite{vaswani2017attention} followed by a two-layer MLP, and add that encoding to the diffusion timestep embedding, which is then projected and added to each residual block in the U-Net.


To enable separate guidance for camera and time, 
we randomly drop out either $\cfgCondTime=(\condTime, \tgtTime)$ or both $\cfgCondTime$ and $\cfgCondImagePose=(\condCamera, \condImage)$ during training, with a probability $7.5\%$ for each case.
We sample from the model using classifier-free guidance as in \cite{watson2024controlling}:
\begin{equation*}
\newcommand{\padvarnothing}{\,\,\varnothing\,}
\begin{aligned}
&\unet\left(\latentT, \tgtCamera, \varnothing, \varnothing\rgt) \\
+ \, \cfgScaleImagePose \cdot &\lft[\unet\lft(\latentT, \tgtCamera, \cfgCondImagePose, \varnothing\rgt) - \unet\lft(\latentT, \tgtCamera, \varnothing, \varnothing\rgt)\right] \\
+ \, \cfgScaleTime \cdot &\left[\unet\lft(\latentT, \tgtCamera, \cfgCondImagePose, \cfgCondTime\rgt) - \unet\lft(\latentT, \tgtCamera, \cfgCondImagePose, \varnothing\rgt)\right],
\end{aligned}
\end{equation*}
where $\unet$ is the denoising network, $\latentT$ are the latents of all target images at diffusion timestep $i$, and $\cfgScaleImagePose$ and $\cfgScaleTime$ are two guidance scale hyperparameters. Intuitively, $\cfgScaleTime$ is to enhance the time alignment of the generated samples, while $\cfgScaleImagePose$ is to encourage consistency with the other conditioning information besides time. 
We initialize the model with the checkpoints from CAT3D, and train with $\ncond=3$ input views and $\ntgt=13$ target views.
For all of our experiments, we set $\cfgScaleImagePose=3.0$ and $\cfgScaleTime=4.5$.
Please refer to the supplementary materials for more model and training details.

\subsection{Dataset Curation}
\label{sec:method_data}

To fully disentangle camera and time controls (such that $\tgtCamera$ only controls camera motion and $\tgtTime$ only controls scene motion), we would ideally train on a large-scale dataset of multi-view videos that capture a dynamic 3D scene from multiple perspectives.
However, this kind of dataset does not yet exist at scale due to the expense of collecting video captures from multiple synchronized cameras~\cite{grauman2024ego,sener2022assembly101,huang2024egoexolearn}.
Rendering such a dataset from synthetic assets is straightforward, but existing synthetic 4D datasets~\cite{greff2022kubric,deitke2023objaverse,zheng2023pointodyssey,raistrick2023infinite} by themselves are insufficiently diverse or realistic to train a model that generalizes to challenging real-world scenes.
Therefore, we carefully curate available datasets for training to maximally cover different combinations of camera and scene motions. See \tabref{data} for an overview of our real and synthetic training datasets, grouped by the characteristics of motions in input and target views.

Fortunately, there are many real-world multi-view images of static scenes~\cite{reizenstein2021common,yu2023mvimgnet,zhou2018stereo,li2019learning} and real-world monocular videos taken at a static viewpoint, each of which corresponds to one of our control signals: camera or time.
We therefore use a mixture of synthetic 4D datasets~\cite{greff2022kubric,deitke2023objaverse}, multi-view image datasets~\cite{reizenstein2021common,yu2023mvimgnet,zhou2018stereo,li2019learning}, and monocular video datasets.
We filter the video dataset to contain only videos of a static viewpoint, in order to prevent the model from confusing time control $\tgtTime$ with camera motion (as we lack camera annotation for these videos).
We perform this filtering by checking if the four corner patches of each video are constant over time (see supplement). 

Until this point, we still do not have real data corresponding to the two cases we particularly care about --- both camera and scene moves in the input views but one of them stays static in the target views.
To address this issue, we perform generative data augmentations~\cite{trevithick2024simvs}.
Specifically, on the one hand, we take samples from CO3D dataset~\cite{reizenstein2021common}, and prompt a video generation model (Lumiere~\cite{bar2024lumiere}) to animate the input frames without moving the camera.
On the other hand, we take samples from the video dataset and run CAT3D~\cite{gao2024cat3d} to generate novel views of the input frames.
While augmented images are often not perfect (\eg, unrealistic deformation, incorrect scene scale), we found them to help the model better learn disentangled camera and time control.

\begin{table}[t]

\setlength{\tabcolsep}{2.5pt}

\resizebox{\columnwidth}{!}{
\Huge
\begin{tabular}{@{}cc|cc|c|c@{}}
\toprule
\multicolumn{2}{c|}{Input views} & \multicolumn{2}{c|}{Target views} & \multirow{3}{*}{Real data} & \multirow{3}{*}{Synthetic data} \\
\cmidrule{1-4}
\makecell{Camera\\motion}   & \makecell{Scene\\motion}  & \makecell{Camera\\motion}   & \makecell{Scene\\motion}   &                            &                                 \\
\midrule
\greencmark          & \redxmark        & \greencmark           & \redxmark         & \makecell{CO3D~\cite{reizenstein2021common}, MVImgNet~\cite{yu2023mvimgnet}\\ Re10K~\cite{zhou2018stereo}, MC4K~\cite{li2019learning}}& \multirow{6}{*}{\makecell{Kubric~\cite{greff2022kubric},\\Objaverse~\cite{deitke2023objaverse}}} \\
\cellcolor{tabgray}\redxmark          & \cellcolor{tabgray}\greencmark        & \cellcolor{tabgray}\redxmark           & \cellcolor{tabgray}\greencmark         & \cellcolor{tabgray}static-view videos              &                            \\
\greencmark          & \greencmark        & \greencmark           & \greencmark         & -                           &                                    \\
\cellcolor{tabgray}\greencmark          & \cellcolor{tabgray}\greencmark        & \cellcolor{tabgray}\greencmark           & \cellcolor{tabgray}\redxmark         & \cellcolor{tabgray}
\makecell{CO3D \\  augmented with Lumiere~\cite{bar2024lumiere}}  &                                   \\
\greencmark          & \greencmark        & \redxmark           & \greencmark         & \makecell{static-view videos\\augmented with CAT3D~\cite{gao2024cat3d}}         &                                   \\
\cellcolor{tabgray}\greencmark          & \cellcolor{tabgray}\greencmark        & \cellcolor{tabgray}\redxmark           & \cellcolor{tabgray}\redxmark         & \cellcolor{tabgray}single image                  &                                   \\
\bottomrule
\end{tabular}
}
\vspace{-0.1in}
\caption{{\bf Our training datasets:} Datasets grouped based on whether each has camera or scene motion in the input or target views. The ``single image'' row corresponds to randomly (with 1\% probability) setting all target views to be one of the input views when drawing samples from all above datasets. For Objaverse, we only use the filtered animated assets~\cite{liang2024diffusion4d}. See \secref{method_data} for details.}
\vspace{-0.2in}
\label{tab:data}
\end{table}

\subsection{Sparse-View Bullet-Time 3D Reconstruction}
\label{sec:method_bullettime}
One application of our model is sparse-view \emph{static} 3D reconstruction in the presence of dynamic scene motion.
Given only a few posed images of a dynamic scene, we want to create a ``bullet-time'' effect by reconstructing a static 3D scene corresponding to the time of one input view.
To achieve this, we follow a similar two-step approach to CAT3D~\cite{gao2024cat3d} --- we first use the diffusion model with an anchored sampling strategy to generate $K$ novel views at some target time, and then run a robust 3D reconstruction pipeline on those images.

Concretely, we first generate $\ntgt$ anchor views by specifying all target times $\tgtTime$ to be the same as one of the inputs $t\in\condTime$.
Then, we split all $K$ target views into batches of $\ntgt$ views and generate each batch by conditioning on the nearest $\ncond$ anchor views.
Finally, we reconstruct the 3D scene using a standard 3DGS model~\cite{kerbl3Dgaussians} with an additional perceptual loss LPIPS~\cite{zhang2018unreasonable}.
See the supplement for details.


\subsection{Generating Consistent Multi-view Videos}
\label{sec:method_sampling}
\begin{figure}
    \centering
    \includegraphics[width=0.99\linewidth]{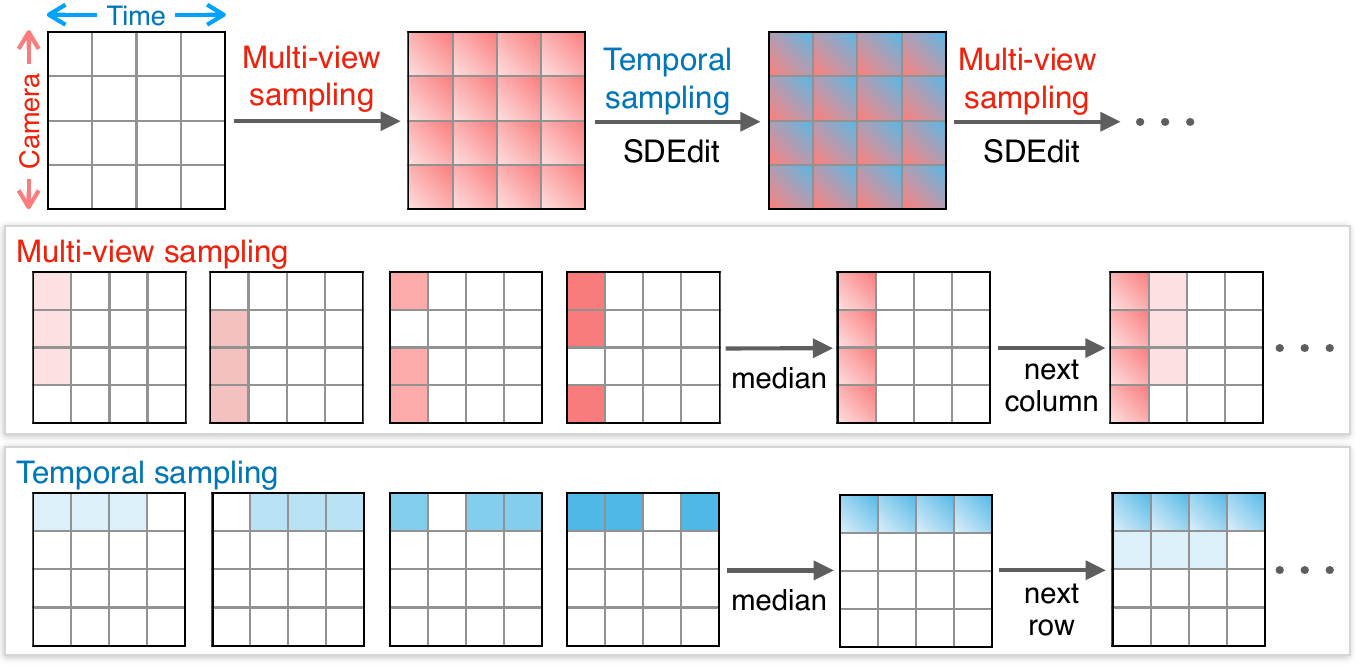}
    \vspace{-0.1in}
    \caption{{\bf Illustrating our alternating sampling strategy:} Given a diffusion model that generates $\ntgt$ output views (here, $\ntgt=3$), we use SDEdit~\cite{meng2021sdedit} to alternate between multi-view and temporal sampling to generate a grid of images at $K$ cameras and $L$ time steps (top, here $K=4$ and $L=4$). In multi-view sampling, we generate each sliding window of size $3$ for each column and take the median of the results (middle). Temporal sampling follows a similar process for rows (bottom). Generations for each column or row can be executed in parallel.}
    \vspace{-0.1in}
    \label{fig:sampling}
\end{figure}

To enable 4D reconstruction, we transform the input monocular video to a multi-view video. 
When there is sufficient camera movement in the monocular input video, we aim to generate multi-view video at a subset of the observed cameras. 
Given an input video of $L$ frames $\Image_{1:L}$ with cameras $\Camera_{1:L}$, we start by picking $K$ camera viewpoints $\{\Camera_{k_i}\}_{i=1}^K\subseteq\Camera_{1:L}$ by farthest point sampling.
Our goal is to generate multi-view videos at these $K$ cameras, \ie, a $K\times L$ grid of images $\gridImage_{K,L}$.
Recall that our model is trained to generate $\ntgt$ frames only, yet usually $KL\gg N$.
To generate such a grid of images, we design a sampling strategy that alternates between \textit{multi-view sampling} that independently generates each column of the image grid and \textit{temporal sampling} that independently generates each row of the image grid. Here we use a restricted setting of our multi-view video diffusion model where the number of conditioning frames $M$ is equal to the number of target frames $N$ + 1. In order to condition on more input frames, we use a fine-tuned version of our trained model with $\ncond=9$ input views and $\ntgt=8$ output views. 
See \figref{sampling} for an illustrative example of this procedure.

\myparagraph{Multi-view sampling}
For each time $\Time$, we generate images $\gridImage_{\cdot, \Time}$ at all $K$ cameras by first generating each sliding window of size $\ntgt$ and then taking the pixel-wise median of the results.
The generation of the $j$-th window is conditioned on $\ntgt$ input frames at corresponding target cameras $\{\Image_{c} \mid c\in \{k_{i \bmod K}\}_{i=j}^{j+\ntgt}\}$ and $1$ input frame at the target time $\Image_t$.
We choose such conditional frames as they provide the most useful information.
Taking the pixel-wise median of multiple windows aggregates information from more conditional frames and also helps reduce the variance.
Results from multi-view sampling are largely multi-view consistent, but are not temporally consistent.

\myparagraph{Temporal sampling}
For each camera $\Camera_{k_i}$, we generate images $\gridImage_{i, \cdot}$ at all $L$ times using a similar sliding window approach as multi-view sampling.
The generation of the $j$-th window is conditioned on $\ntgt$ input frames at corresponding target timestamps $\{\Image_{t \bmod L}\}_{t=j}^{j+\ntgt}$ and $1$ input frame at target camera $\Image_{k_i}$.
Results from temporal sampling are largely temporally consistent, but are not multi-view consistent.

\myparagraph{Alternating strategy}
To achieve both multi-view and temporal consistency, we alternate between multi-view sampling and temporal sampling with SDEdit~\cite{meng2021sdedit} from the previous iteration and use 25 DDIM steps for sampling from random noise and a reduces
number of steps when initializing from lower noise levels. 
We run three iterations of sampling: (1) multi-view sampling from random noise,  (2) temporal sampling initialized from the previous iteration at noise level $16/25$ (3) multi-view sampling initialized from the previous iteration at noise level $8/25$. 
Multi-view videos generated with this strategy are sufficiently consistent for accurate 4D reconstruction, provided the input video has enough coverage of the scene (recall that the $K$ cameras are picked from the input cameras $\Camera_{1:L}$).

\myparagraph{Stationary videos}\label{par:static} For input videos with little to no camera movement, we must generate images from novel viewpoints to sufficiently constrain 4D reconstruction.
To do this, we first generate $K$ novel views at $t=0$, add these generated frames to the set of input frames, and then run the previously described alternating sampling strategy.

\myparagraph{Dense view sampling}
To further increase the coverage of our generated multi-view videos, for each timestep we condition on the $K$ generated views and generate $K'$ more views using the nearest-anchoring strategy presented in \secref{method_bullettime}.
For our experiments, we set $K=13$ and $K'=128$.
We use MonST3R~\cite{zhang2024monst3r} to get camera parameters for unposed input videos.
More details (\eg, sampled camera trajectories) are described in the supplement.


\begin{figure}[t!]
    \centering
    \includegraphics[width=0.99\linewidth]{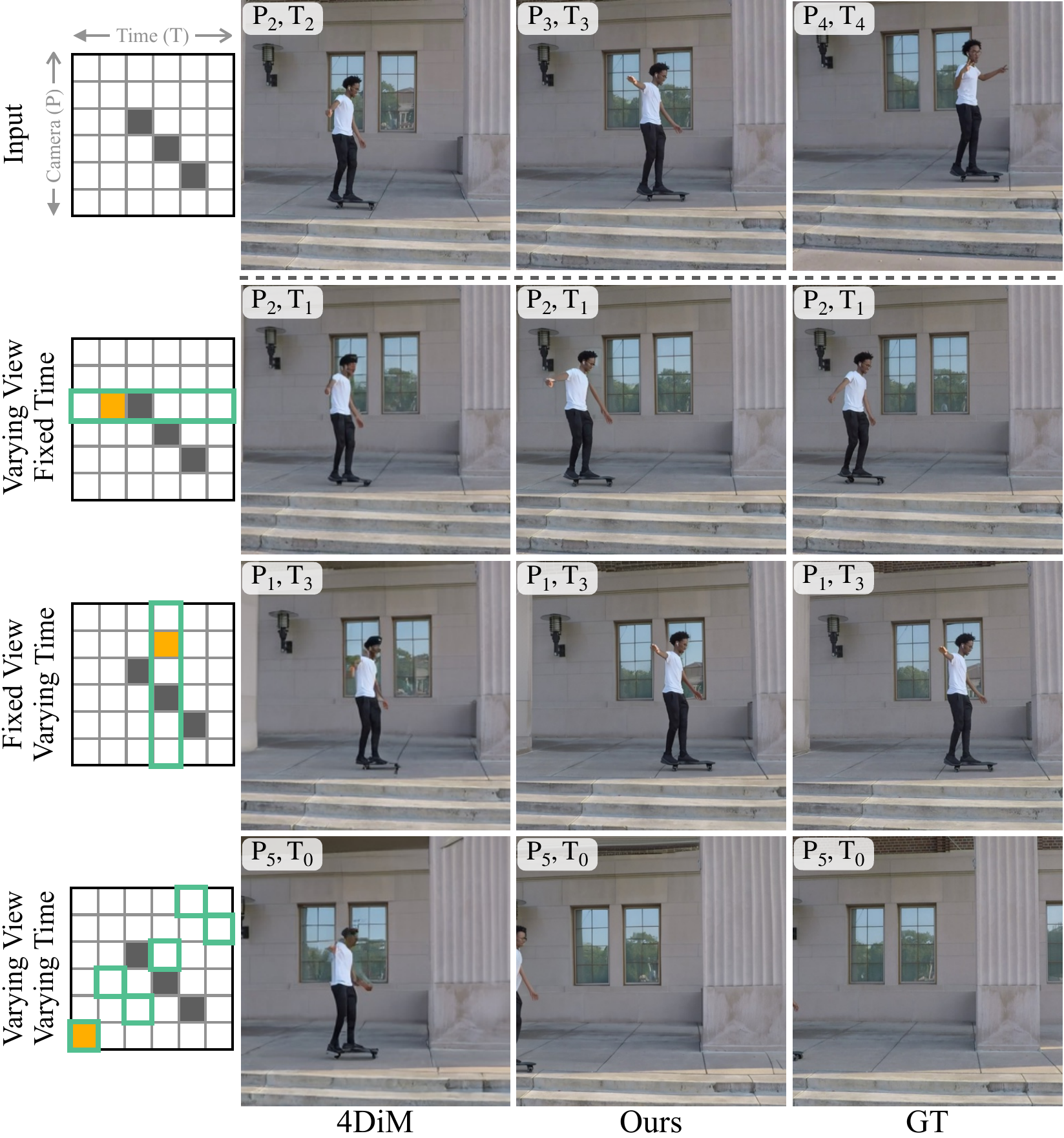}
    \vspace{-0.1in}
    \caption{\textbf{Qualitative comparison, disentangled control:} The camera-time grid on the left shows the positions of three input images (gray cells, images visualized in top row) and output images (green) in three different target sampling settings. One frame from each setting (orange cell) is visualized in each row, comparing our model with 4DiM~\cite{watson2024controlling} and ground truth.}
    \vspace{-0.1in}
    \label{fig:cameratime_compare}
\end{figure}

\subsection{4D Reconstruction}
\label{sec:method_4d}
The alternating sampling method (\secref{method_sampling}) generates multi-view videos that are sufficiently consistent to be used in existing 4D reconstruction pipelines.
We build on 4D-GS~\citep{wu20244d}, which represents a dynamic 3D scene as a set of canonical-space 3D Gaussians that are moved by a deformation field parameterized with K-Planes~\cite{fridovich2023k}.
See the supplement for details.
\section{Experiments}
\label{sec:experiments}

\begin{figure}[t!]
    \centering
    \includegraphics[width=0.99\linewidth]{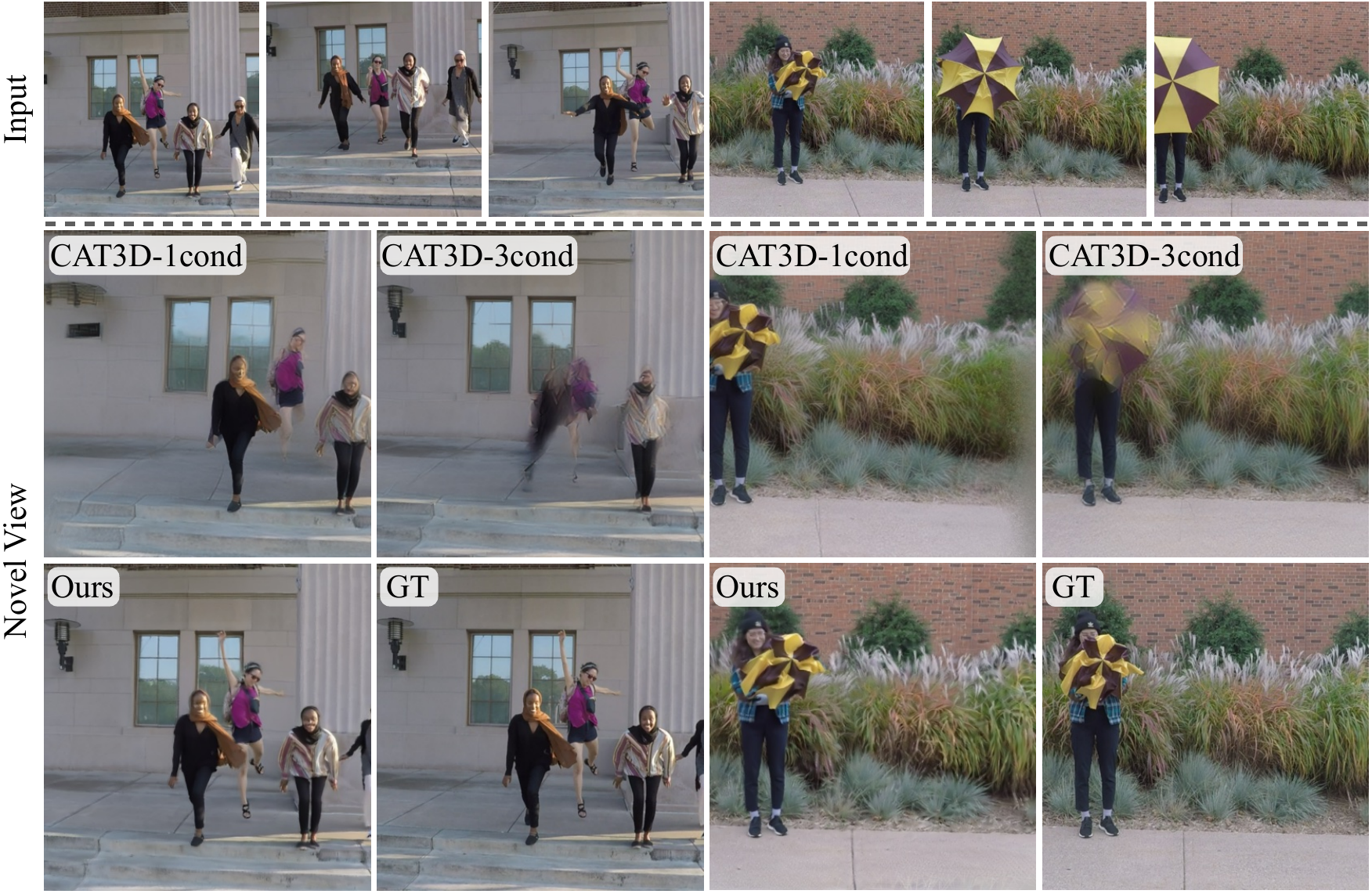}
    \vspace{-0.1in}
    \caption{{\bf Qualitative comparison, sparse-view ``bullet-time'' 3D reconstruction:} The three input images are shown on the top, where the first one is the target bullet-time frame.}
    \label{fig:bullettime_3drec_compare}
\end{figure}

\begin{figure}[t!]
    \centering
    \includegraphics[width=0.95\linewidth]{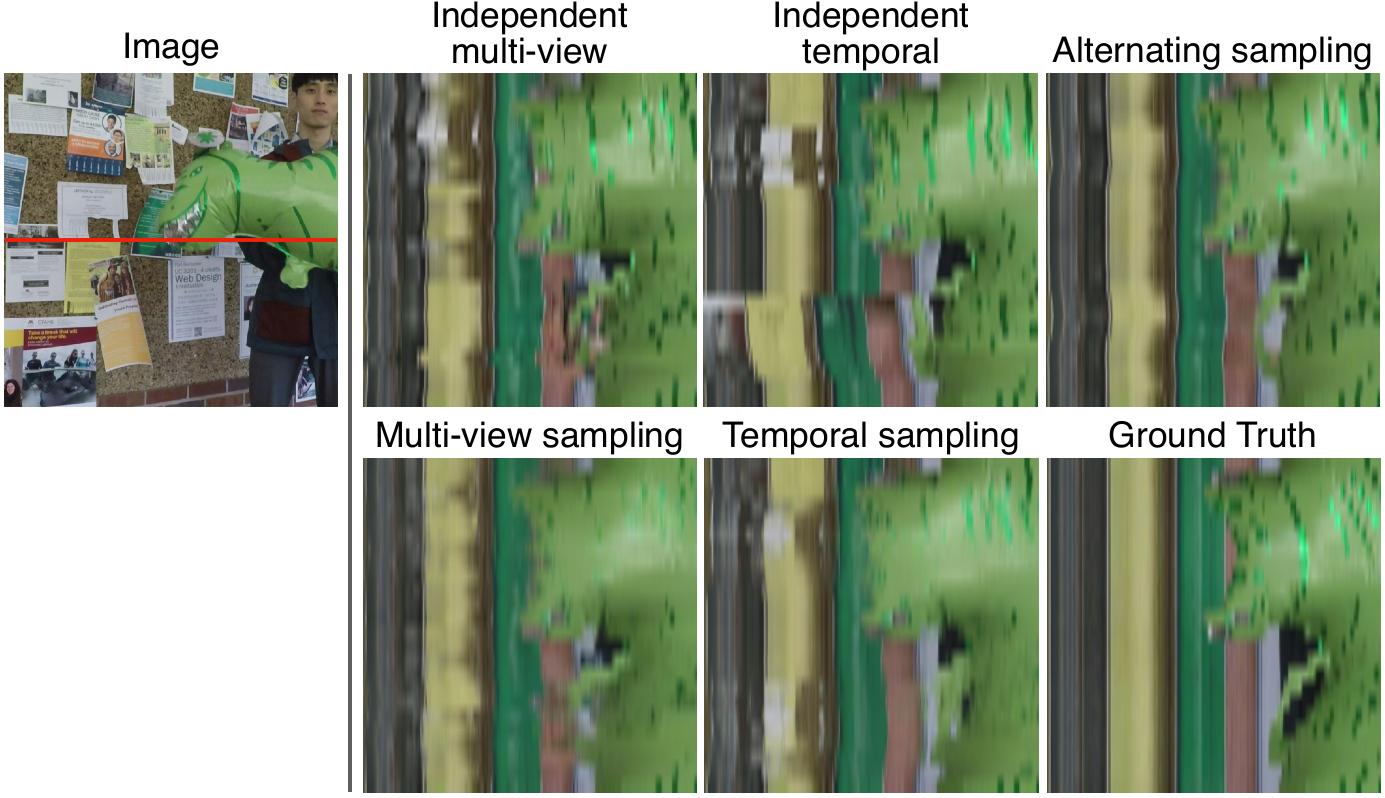}
    \vspace{-0.1in}
    \caption{\textbf{Qualitative comparison, sampling strategies:} A comparison of different sampling strategies using space-time slices, where the vertical axis represents time and the horizontal axis shows a spatial slice of the image (red line). Our alternating sampling strategy best matches the ground truth motion.} 
    \vspace{-0.1in}
    \label{fig:ablation}
\end{figure}

\myparagraph{Separate Control over Camera and Time} 
We first examine our model's ability to separately control camera viewpoint and scene dynamics.
Given $3$ input images, we generate three types of output sequences using our diffusion model:
1) fixed viewpoint and varying time,
2) varying viewpoint and fixed time, and
3) varying viewpoint and varying time.
We evaluate on the NSFF dataset~\cite{li2021neural} (a subset of the Nvidia Dynamic Scenes Dataset~\cite{yoon2020novel}), which provides ground truth frames from $12$ synchronized cameras in eight real-world dynamic scenes.
For each scene, we test on $8$ different sets of input images that are randomly sampled from the dataset.
We compare against 4DiM~\cite{watson2024controlling}, a recent method that trains a cascaded diffusion model to synthesize images under specified viewpoints and timestamps.
Qualitative and quantitative results are shown in \figref{cameratime_compare} and \tabref{cameratime_compare}, respectively.
We observe that 4DiM conflates control over camera and scene motion: even when the model is instructed to only change the camera viewpoint, dynamic objects still move.
In contrast, our model exhibits decoupled control over camera and scene motion control, in addition to generating overall higher quality images.

\begin{table}[t]

\setlength{\tabcolsep}{2.2pt}
\renewcommand{\arraystretch}{1.2}

\resizebox{\columnwidth}{!}{
\begin{tabular}{@{}l|c@{\,}c@{\,}c|c@{\,}c@{\,}c|c@{\,}c@{\,}c@{}}
\toprule
\multirow{3}{*}{Method} & \multicolumn{3}{c|}{\makecell{Fixed Viewpoint\\Varying Time}}    & \multicolumn{3}{c|}{\makecell{Varying Viewpoint\\Fixed Time}}    & \multicolumn{3}{c}{\makecell{Varying Viewpoint\\Varying Time}}         \\
\cmidrule{2-10}
     & {PSNR} & {SSIM} & {LPIPS} & {PSNR} & {SSIM} & {LPIPS} & {PSNR} & {SSIM} & {LPIPS} \\
\midrule
4DiM\cite{watson2024controlling} &                      19.77 &                      0.540 &                      0.195 &                      18.81 &                      0.428 &                      0.219 &                      17.28 &                      0.378 &                      0.256 \\
Ours                              &  \cellcolor{tabfirst}21.97 &  \cellcolor{tabfirst}0.683 &  \cellcolor{tabfirst}0.121 &  \cellcolor{tabfirst}21.68 &  \cellcolor{tabfirst}0.588 &  \cellcolor{tabfirst}0.105 &  \cellcolor{tabfirst}19.73 &  \cellcolor{tabfirst}0.533 &  \cellcolor{tabfirst}0.155 \\

\bottomrule
\end{tabular}
}
\vspace{-0.1in}
\caption{\textbf{Quantitative comparison, disentangled control:} We compare with 4DiM~\cite{watson2024controlling} on the NSFF dataset~\cite{li2021neural}, evaluating how well the time and viewpoint can be independently manipulated.}
\label{tab:cameratime_compare}

\end{table}

\begin{table}[t]
\centering
\small
\begin{tabular}{lccc}
\toprule
Method        & PSNR$\uparrow$  & SSIM$\uparrow$  & LPIPS $\downarrow$ \\
\midrule
CAT3D-1cond~\cite{gao2024cat3d} &                      15.33 &                      0.379 &                      0.527 \\
CAT3D-3cond~\cite{gao2024cat3d} &                      20.19 &                      0.568 &                      0.258 \\
Ours                            &  \cellcolor{tabfirst}20.79 &  \cellcolor{tabfirst}0.576 &  \cellcolor{tabfirst}0.160 \\
\bottomrule
\end{tabular}
\vspace{-0.1in}
\caption{\textbf{Quantitative comparison, sparse-view ``bullet-time":} We compare with CAT3D~\cite{gao2024cat3d} on the NSFF dataset~\cite{li2021neural}, evaluating the ability to reconstruct a consistent static 3D scene from input images containing scene motion.}
\label{tab:bullettime_3drec_compare}

\end{table}

\begin{figure*}[t!]
    \centering
    \includegraphics[width=0.99\linewidth]{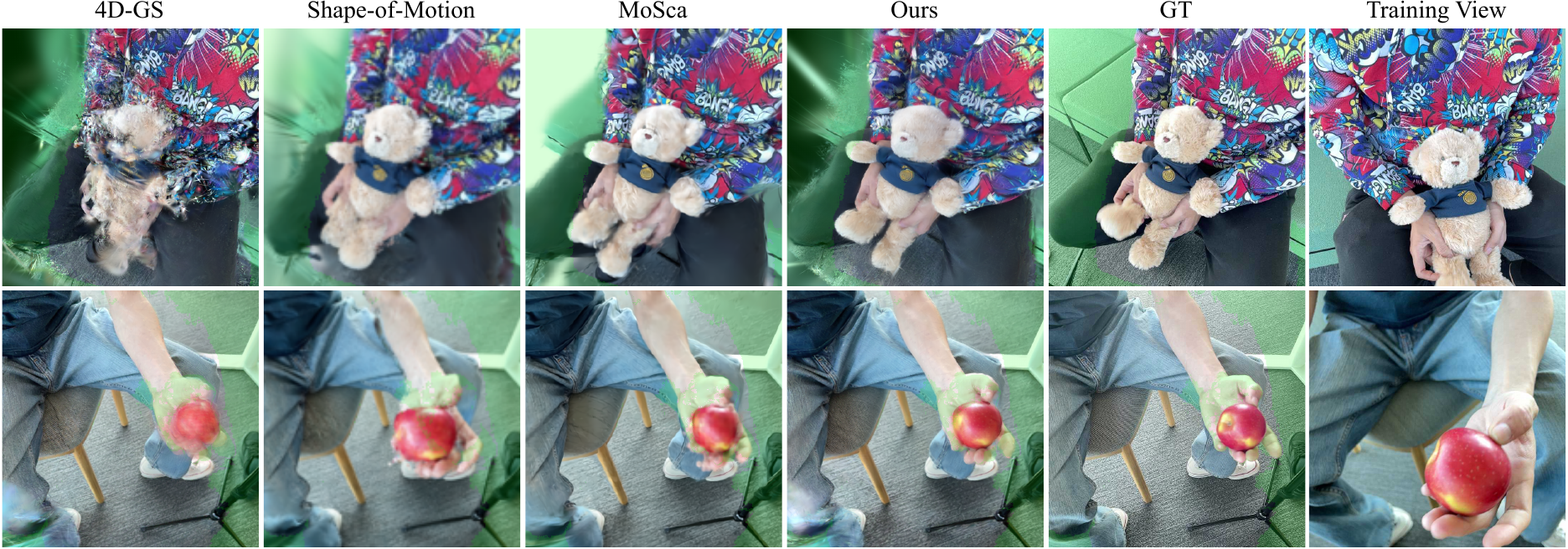}
    \vspace{-0.1in}
    \caption{\textbf{Qualitative comparison, 4D reconstruction:} We compare 4D reconstructions on the DyCheck dataset~\cite{gao2022monocular}. The rightmost column shows the input frame, at the same timestamp as the desired target image. Evaluation-excluded co-visibility masks are highlighted green. The visual improvement of our rendering over 4D-GS~\cite{wu20244d} (which our system leverages) demonstrates the value of our proposed multi-view video model. Renderings from Shape-of-Motion~\cite{wang2024shape} and MoSca~\cite{lei2024mosca} were graciously provided by the authors.}
    \vspace{-0.1in}
    \label{fig:4drec_compare}
\end{figure*}

\myparagraph{Sparse-View Bullet-Time 3D Reconstruction} %
\label{sec:exp_3drec}
On this task, we again evaluate on the NSFF dataset~\cite{li2021neural} with $3$ input images. For each scene, we test on 4 different sets of input images randomly sampled from the captured sequence. 
We compare against two versions of CAT3D~\cite{gao2024cat3d}, with either all $3$ input images (``CAT3D-3cond'') or only one input image at the target time (``CAT3D-1cond'').
Qualitative and quantitative comparison results are presented in \figref{bullettime_3drec_compare} and \tabref{bullettime_3drec_compare}, respectively.
CAT3D-3cond is unable to resolve inconsistencies in the inputs, resulting in blurry novel view renderings, especially in dynamic parts of the scene.
CAT3D-1cond, which is given only one image at the target time, does not need to resolve any dynamic inconsistencies, but it is unable to leverage available information in other input images to determine the global scene scale and correct scene content.
Our multi-view video diffusion model reconstructs the 3D scene at the target time most accurately.

\myparagraph{4D Reconstruction} \label{sec:exp_4drec}
We evaluate the task of dynamic scene reconstruction from monocular video on the DyCheck dataset~\cite{gao2022monocular}, which contains video sequences with 200-500 frames of challenging real world scene motion.
Following \cite{wang2024shape}, we test on $5$ scenes that have two synchronized static cameras for novel-view synthesis evaluation.
In \tabref{4drec_compare} and \figref{4drec_compare}, we compare to the baseline 4D-GS~\cite{wu20244d} (without using our generated images) and state-of-the-art dynamic reconstruction methods Shape-of-Motion~\cite{wang2024shape} and MoSca~\cite{lei2024mosca}.
Our method greatly improves the reconstruction quality over the baseline 4D-GS representation that we used.
Our results are roughly on par with those of Shape-of-Motion and MoSca, but those models use additional supervisory signals (\eg, depth maps, 2D tracks, user-provided foreground masks).
We use no additional signals and supervise with only straightforward photometric losses.

\begin{table}[t]
\resizebox{\columnwidth}{!}{
\begin{tabular}{lccc}
\toprule
Method             & mPSNR$\uparrow$  & mSSIM$\uparrow$  & mLPIPS $\downarrow$ \\
\midrule
4D-GS~\cite{wu20244d}     &  \cellcolor{tabthird}16.54 &  \cellcolor{tabthird}0.594 & \cellcolor{tabsecond}0.347 \\
Shape-of-Motion~\cite{wang2024shape} & \cellcolor{tabsecond}16.72 &  \cellcolor{tabfirst}0.630 &  \cellcolor{tabthird}0.450 \\
Ours                                 &  \cellcolor{tabfirst}17.39 & \cellcolor{tabsecond}0.607 &  \cellcolor{tabfirst}0.341 \\
\midrule
MoSca~\cite{lei2024mosca}$^\dagger$ &  \cellcolor{tabfirst}19.54 &  \cellcolor{tabfirst}0.738 & \cellcolor{tabsecond}0.244 \\
Ours$^\dagger$                      & \cellcolor{tabsecond}18.24 & \cellcolor{tabsecond}0.666 &  \cellcolor{tabfirst}0.227 \\
\bottomrule
\end{tabular}
}
\vspace{-0.1in}
\caption{\textbf{Quantitative comparison, 4D reconstruction:} Following prior work ~\cite{gao2022monocular,wang2024shape,lei2024mosca}, we report co-visibility masked image metrics on the DyCheck dataset~\cite{gao2022monocular}.  $^\dagger$ indicates methods trained on images at half the original resolution.}
\label{tab:4drec_compare}
\end{table}

\myparagraph{4D Creation} 
In addition to reconstructing of real, captured dynamic 3D scenes, CAT4D can also be used to create 4D scenes of generated content, by first using a text-to-video or image-to-video model~\cite{genmo,kling,sora,pika,dreammachine,veo2024} to create a monocular video sequence.    
Unlike real captures, these generated videos sometimes contain inconsistent scene elements that defy physical 3D constraints and typically lack the camera motion that most state-of-the-art reconstruction methods require.
\figref{4dgen_gallery} shows a collection of these results, but we strongly recommend viewing the supplementary video instead.

\begin{table}
\centering
\resizebox{\columnwidth}{!}{
\begin{tabular}{lccc}
\toprule
Sampling strategy & PSNR $\uparrow$ & SSIM $\uparrow$ & LPIPS $\downarrow$ \\
\midrule

Independent multi-view &                      20.27 &                      0.525 &  \cellcolor{tabthird}0.136 \\
Independent temporal   &                      21.63 &  \cellcolor{tabthird}0.615 & \cellcolor{tabsecond}0.130 \\
Multi-view sampling     & \cellcolor{tabsecond}22.34 &                      0.609 &                      0.217 \\
Temporal sampling       &  \cellcolor{tabfirst}23.36 &  \cellcolor{tabfirst}0.681 &                      0.145 \\
Alternating sampling     &  \cellcolor{tabthird}22.15 & \cellcolor{tabsecond}0.633 &  \cellcolor{tabfirst}0.108 \\
\bottomrule
\end{tabular}
}
\vspace{-0.1in}
\caption{\textbf{Quantitative comparison, sampling strategies:} We compare our sampling strategy to four simpler variants by comparing to ground-truth images from the NSFF dataset~\cite{li2021neural}.}
\vspace{-0.1in}
\label{tab:ablation}
\end{table}

\paragraph{Ablation Study} 
The multi-view and temporal consistency of our samples is essential for high-quality 4D creation.
We compare our alternating sampling strategy (\secref{method_sampling}) to four simpler options:
1) multi-view sampling only,
2) temporal sampling only,
3) independent multi-view sampling (\ie, without the sliding-window overlap), and
4) independent temporal sampling (\ie, without the sliding-window overlap).
We evaluate different strategies on the NSFF dataset~\cite{li2021neural} for generating multi-view videos.
The results in \tabref{ablation} and \figref{ablation} shows that our alternating sampling strategy improves sample quality.
We also present an ablation study of our training datasets in the supplement.


\section{Discussion}
We present CAT4D, an approach for creating 4D scenes from captured or generated monocular videos. Our multi-view video diffusion model transforms monocular inputs into consistent multi-view videos, enabling reconstruction as deformable 3D Gaussians. 

CAT4D has several limitations: the diffusion model struggles with temporal extrapolation beyond input frames and cannot fully disentangle camera viewpoint from temporal progression, especially in challenging cases where dynamic objects become occluded. The alternating sampling strategy we propose is effective at increasing the number of frames we can generate, but training larger-scale multi-view video models that can accomplish this directly is an exciting future area. Additionally, while our generated 4D scenes appear plausible from novel viewpoints, the recovered 3D motion fields may not be physically accurate. Incorporating supervision signals like depth or motion estimates could improve the quality of our results but would reduce the applicability of our method to dense video captures.

\myparagraph{Acknowledgments}
We would like to thank Arthur Brussee, Philipp Henzler, Daniel Watson, Jiahui Lei, Hang Gao, Qianqian Wang, Stan Szymanowicz, Jiapeng Tang, Hadi Alzayer and Angjoo Kanazawa for their valuable contributions.
We also extend our gratitude to Shlomi Fruchter, Kevin Murphy, Mohammad Babaeizadeh, Han Zhang and Amir Hertz for training the base text-to-image latent diffusion model.

{
    \small
    \bibliographystyle{ieeenat_fullname}
    \bibliography{main}
}

\appendix
\clearpage
\setcounter{page}{1}
\maketitlesupplementary

\section{Method Details}

\myparagraph{Diffusion Model}
We initialize our model from CAT3D~\cite{gao2024cat3d}, with the additional MLP layers in the new timestamp embedding randomly initialized.
All timestamps $\Time \in \condTime \cup \tgtTime$ (of each set of $\ncond+\ntgt$ frames) are normalized within range $[0, 1]$ and are relative to the first timestamp $\condTime_0$.
We fine-tune the full latent diffusion model (\ie, the denoising U-Net) with $\ncond=3$ input views and $\ntgt=13$ target views for $2.0$M iterations with a batch size of $128$ and a learning
rate of $5\times10^{-5}$.
For the 4D reconstruction application, in order to condition on more input frames, we further fine-tune the model with $\ncond=9$ input views and $\ntgt=8$ target views for $20$K iterations.

\myparagraph{Sampling}
For all of our experiments, we use DDIM~\cite{song2020denoising} with $25$ sampling steps and classifier-free guidance weights $s_1=3.0, s_2=4.5$.
Our alternating sampling strategy takes about $1$ minute to generate all $K'=128$ views for each timestamp, when executed in parallel on 16 A100 GPUs.
We note that CAT3D originally use $50$ DDIM steps, yet for our model we found  that $25$ steps work just as well---and using fewer steps reduces the runtime of our sampling strategy.

\begin{figure*}
    \centering
    \includegraphics[width=0.99\linewidth]{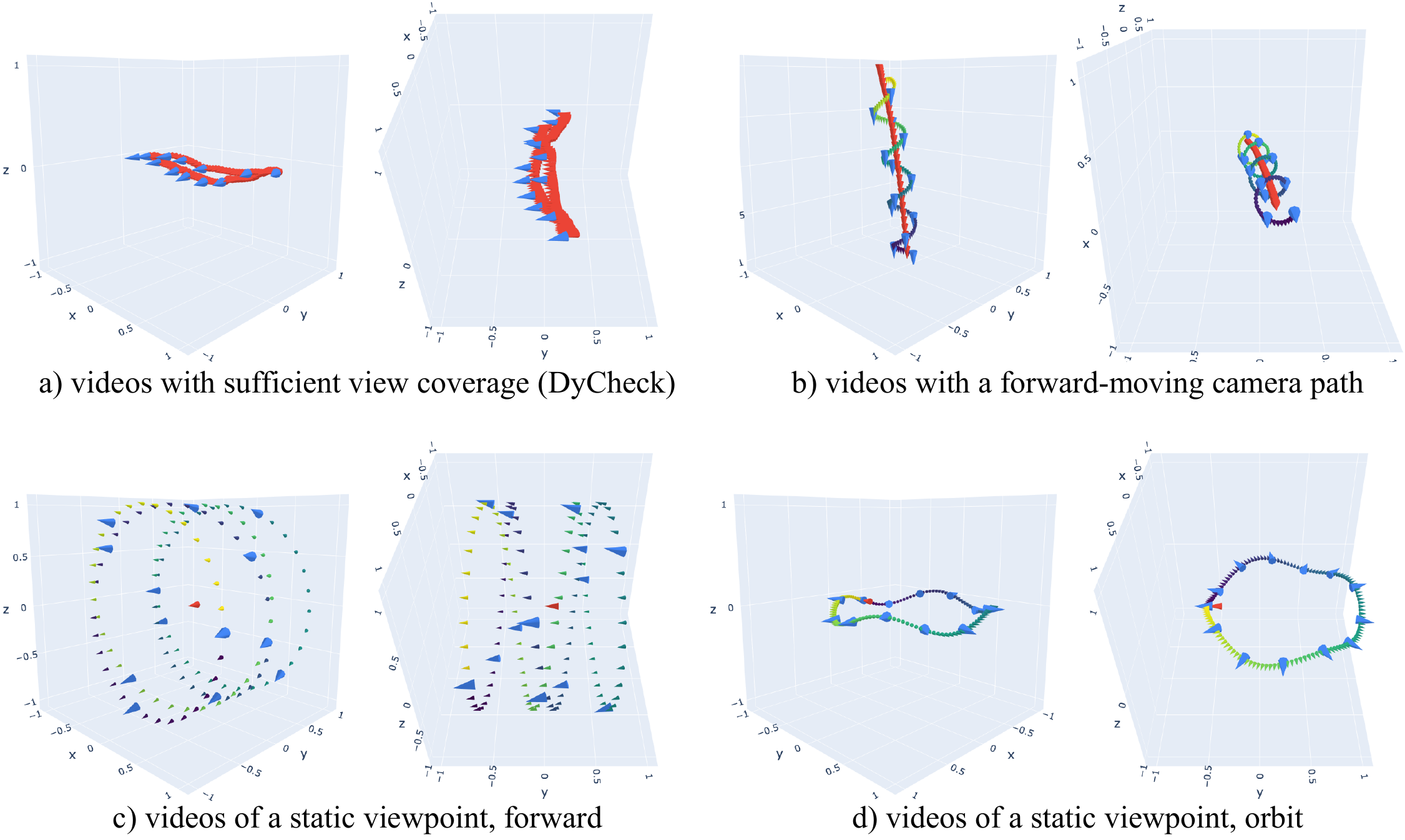}
    \vspace{-0.05in}
    \caption{Camera trajectories (where we generate novel views) for different types of input videos. Within each panel, we show the trajectories from two different viewpoints. The input views are colored red, and the anchoring sample views are colored blue with the remaining sample views are colored by their index. For videos with sufficient view coverage (a), we only generate anchor views picked from the input camera trajectory.}
    \label{fig:cameras}
\end{figure*}

\myparagraph{Camera Trajectory Selection}
The choice of camera trajectories where we generate novel views has a large impact on the quality of 4D creation.
In principle, the camera trajectories should cover the viewpoints where we want to render the scene after reconstruction.
We design the novel-view camera trajectory based on the camera trajectory of the input video:
\begin{itemize}
    \item For input videos with sufficient view coverage (\eg, videos from DyCheck~\cite{gao2022monocular} whose cameras are centered around a focus point), we simply sample views on the input camera trajectory.
    \item For input videos with a forward-moving camera trajectory, we sample novel views from a spiral path around the input camera trajectory.
    \item For input videos with little or no camera movement, we sample novel views from either a spiral path that moves into and out of the scene or a orbit path that spins around the central object. For each example, we run both, and select the one which is most appropriate for the given scene.
\end{itemize}
See \figref{cameras} for an illustration of different types of camera trajectories.

\myparagraph{Sparse-View Bullet-Time 3D Reconstruction}
The conditional times $\condTime$ should in principal be the actual timestamps of the input frames, but they may not be known for unstructured in-the-wild datasets.
We found that in practice the model works well if we just set the timestamps for the bullet-time frame as 0 and other frames as 1.

\myparagraph{4D Reconstruction}
We build our reconstruction pipeline on top of 4D-GS~\cite{wu20244d} with several extensions.
We use a combination of L1, DSSIM and LPIPS for the photometric reconstruction loss, with weighting factors $0.8$, $0.2$ and $0.4$ respectively, and keep all regularization terms from ~\cite{wu20244d} as is.
We set the 3D Gaussians' densification threshold (magnitude of view-space position gradients) to $0.0004$, and use a batch size of $4$ (images).
We initialize the 3D Gaussians with points from SfM~\cite{snavely2006photo} or MonST3R~\cite{zhang2024monst3r}.
When SfM or MonST3R points are not available, \eg, input videos of a static viewpoint, we use uniformly random points for initialization.
We first optimize only the canonical-space 3D Gaussians with all generated images at $t=0$ for $2000$ iterations,
then jointly optimize both the 3D Gaussians and the deformation field with images at all timestamps for $18000$ iterations.
After the first $2000$ iterations, we linearly anneal the multiplier of reconstruction loss for our generated images from $1.0$ to $0.5$ while keeping the multiplier for real input images fixed to $1.0$.
The optimization takes about $25$ minutes on a single A100 GPU.

\section{Datasets Details}
For Objaverse~\cite{deitke2023objaverse}, we use only the animated assets filtered by \cite{liang2024diffusion4d} (around $42$k in total).
We render each asset under $4$ different lighting conditions ( randomly sampled environment maps).
For each lighting condition, we render synchronized videos of $8$ frames at $8$ evenly spaced viewpoints on the $360^{\circ}$ orbit path.
For Kubric~\cite{greff2022kubric}, we randomly generate $4$k scenes using their generator.
For each scene, we render synchronized videos of $8$ frames at $8$ viewpoints evenly spaced on a smooth camera path that is randomly sampled on the upper hemisphere.
When drawing samples from these two synthetic 4D datasets, we sample the input and target views according to different combinations in \tabref{data} with equal probability.

For data samples from all multi-view image datasets~\cite{reizenstein2021common,yu2023mvimgnet,zhou2018stereo,li2019learning}, we set all timestamps $\condTime \cup \tgtTime$ to zero.

For our video dataset, we filter it to contain only videos of at static viewpoint.
We perform this filtering by checking if the four corner patches (size $10\times 10$) of each video are nearly constant over time.
Concretely, we compute for each corner patch the L2 distance between consecutive frames (averaged over time), and then check if the maximum of the four is smaller than $0.05$.
While this simple strategy sometimes yields false positives, it's sufficiently effective and can be run on the fly.
For samples from this dataset, we randomly shuffle the order of frames and set all camera parameters $\condCamera\cup\tgtCamera$ to be the same with a central principal point and a random focal length sampled from $[0.8\cdot512, 1.2\cdot512]$.

We use Lumiere~\cite{bar2024lumiere} to augment the CO3D dataset~\cite{reizenstein2021common} ($4$-th row in \tabref{data}) following SimVS~\cite{trevithick2024simvs}.
For each sampled sequence ($\condImage, \condCamera, \condTime,  \tgtImage, \tgtCamera, \tgtTime$) from CO3D, we animate each of the $\ncond$ input images (except the first one) using Lumiere~\cite{bar2024lumiere}, resulting in $\ncond-1$ videos $\{V^i_L\}_{i=1}^{M-1}$ of length $L$.
Then we randomly sample one frame (index $k_i$) from each video, and treat them as pseudo ground truth of original input images at another timestamps, \ie $\condImage \gets \{\condImage_0\} \cup \{V_{k_i}^i\}_{i=1}^{M-1}$ and $\condTime \gets \{\condTime_0\} \cup \{\frac{k_i}{L - 1}\}_{i=1}^{M-1}$.
We obtain around $24$k sequences in total with this augmentation.

We use CAT3D~\cite{gao2024cat3d} to augment our static-view video dataset as follows ($5$-th row in \tabref{data}).
For each sampled sequence ($\condImage, \condCamera, \condTime,  \tgtImage, \tgtCamera, \tgtTime$) from the video dataset, we use CAT3D to generate $7$ novel views for each of the $\ncond$ input images (except the first one), resulting in $\ncond-1$ image sets $\{V^i_L\}_{i=1}^{M-1} (L=7)$ at viewpoints $\{\tilde P^i\}_{i=1}^{M-1}$.
Then we randomly sample one frame (index $k_i$) from each image set, and treat them as pseudo ground truth of original input images at another viewpoints, \ie $\condImage \gets \{\condImage_0\} \cup \{V_{k_i}^i\}_{i=1}^{M-1}$ and $\condCamera \gets \{\condCamera_0\} \cup \{\tilde P^i_{k_i}\}_{i=1}^{M-1}$.
We obtain around $160$k sequences in total with this augmentation.

We mix all the datasets (Objaverse~\cite{deitke2023objaverse}, Kubric~\cite{greff2022kubric}, Re10K~\cite{zhou2018stereo}, MVImgNet~\cite{yu2023mvimgnet}, CO3D~\cite{reizenstein2021common}, MQ4K~\cite{li2019learning}, static-view video data, augmented CO3D and augmented static-view video data) with weights $2.5$, $2.5$, $1.0$, $1.0$, $1.0$, $1.0$, $5.0$, $1.0$ and $1.0$, respectively.

\section{Baselines Details}
For the evaluation of sparse-view bullet-time 3D reconstruction, we run CAT3D baselines with one or three input images.
For CAT3D-3cond, we use the same camera trajectory as ours (\figref{cameras} (c)) for generating novel views.
For CAT3D-1cond, we use their default camera trajectory (a forward-facing spiral path similar to \figref{cameras} (c)) for generating novel views, and manually adjust the global scene scale such that it roughly matches the actual scene scale of the dataset.

For the ablation study of different sampling strategies, we evaluate the quality of the generated multi-view videos on the NSFF dataset~\cite{li2021neural}.
As in the prior work~\cite{li2021neural,yoon2020novel}, we simulate a moving monocular camera by extracting images from each of the $12$ camera viewpoints at different timestamps ($24$ in total) and compare the generated multi-view videos at all $12$ viewpoints to the ground truth.

\section{Ablation Study of Training Data}

\begin{figure}[t]
    \centering
    \includegraphics[width=0.99\linewidth]{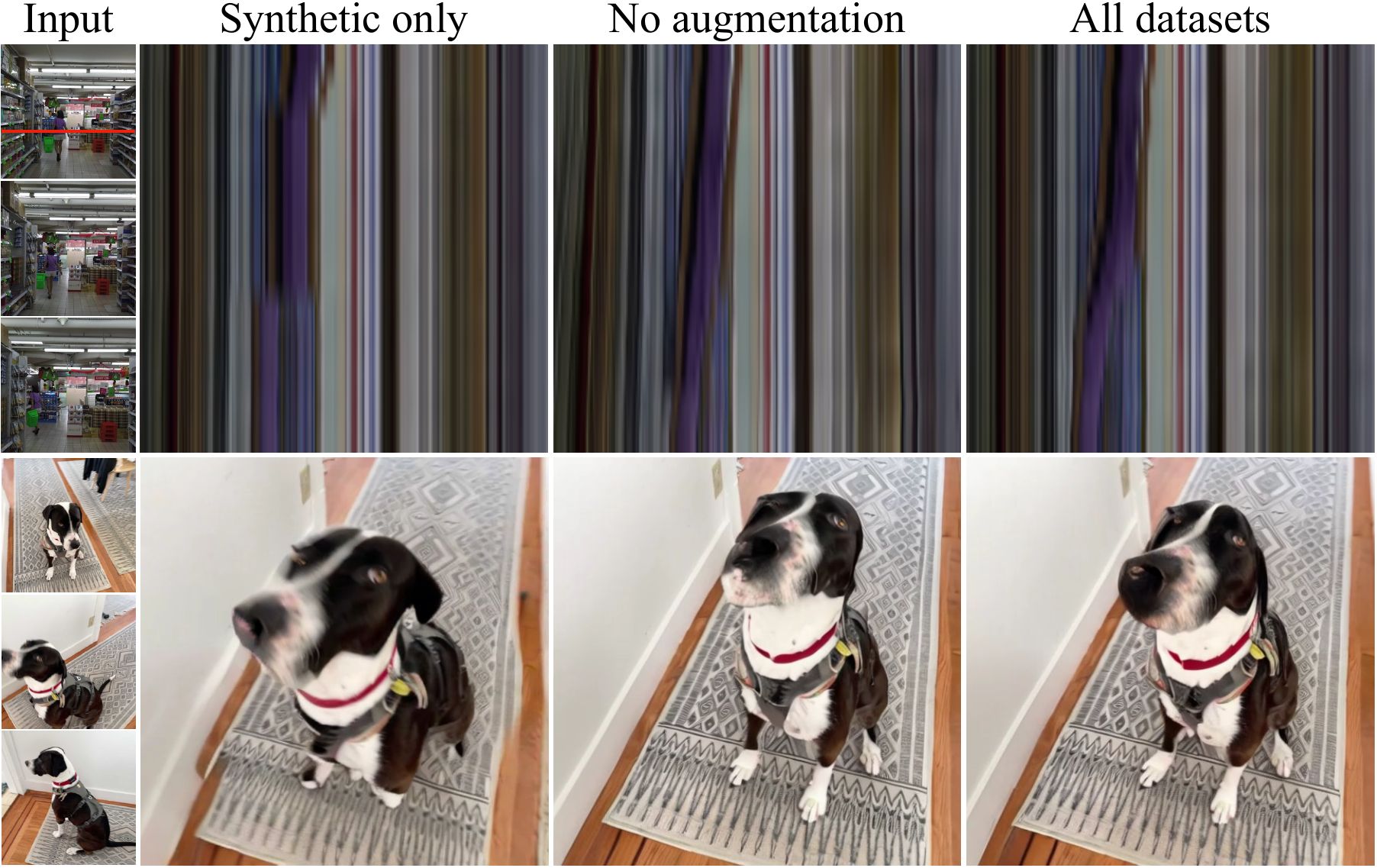}
    \vspace{-0.05in}
    \caption{A comparison of models trained with different datasets on in-the-wild input images. The three input images are shown on the left-most column. Top: space-time slices of generated videos of ``fixed viewpoint, varying time''. Pixels of static background should be straight vertical lines on the slices and pixels of dynamic object should be smooth curves on the slices. Bottom: one frame of generated videos of ``varying viewpoint, fixed time''.}
    \label{fig:ablation_data}
\end{figure}

\begin{table}[ht]

\setlength{\tabcolsep}{2.2pt}
\renewcommand{\arraystretch}{1.2}

\resizebox{\columnwidth}{!}{
\begin{tabular}{@{}l|c@{\,}c@{\,}c|c@{\,}c@{\,}c|c@{\,}c@{\,}c@{}}
\toprule
\multirow{3}{*}{Training Data} & \multicolumn{3}{c|}{\makecell{Fixed Viewpoint\\Varying Time}}    & \multicolumn{3}{c|}{\makecell{Varying Viewpoint\\Fixed Time}}    & \multicolumn{3}{c}{\makecell{Varying Viewpoint\\Varying Time}}         \\
\cmidrule{2-10}
     & {PSNR} & {SSIM} & {LPIPS} & {PSNR} & {SSIM} & {LPIPS} & {PSNR} & {SSIM} & {LPIPS} \\
\midrule
Synthetic only & \cellcolor{tabsecond}22.19 & \cellcolor{tabsecond}0.745 & \cellcolor{tabsecond}0.123 &                      21.41 &                      0.547 &                      0.123 & \cellcolor{tabsecond}19.50 & \cellcolor{tabsecond}0.523 &                      0.173 \\
No augmentation &                      20.84 &                      0.596 &                      0.135 &  \cellcolor{tabfirst}22.03 &  \cellcolor{tabfirst}0.602 &  \cellcolor{tabfirst}0.104 &                      19.41 &                      0.519 & \cellcolor{tabsecond}0.160 \\
All datasets          &  \cellcolor{tabfirst}22.49 &  \cellcolor{tabfirst}0.749 &  \cellcolor{tabfirst}0.110 & \cellcolor{tabsecond}21.86 & \cellcolor{tabsecond}0.599 & \cellcolor{tabsecond}0.105 &  \cellcolor{tabfirst}19.74 &  \cellcolor{tabfirst}0.546 &  \cellcolor{tabfirst}0.152 \\

\bottomrule
\end{tabular}
}
\vspace{-0.1in}
\caption{A ablation study of training data, evaluated on the NSFF dataset~\cite{li2021neural}. All datasets: using all of our training datasets. No augmentation: dropping the two augmented datasets (CO3D augmented with Lumiere and static-view video data augmented with CAT3D). Synthetic only: dropping all real-world datasets and using only synthetic 4D data (Kubric and Objaverse).}
\label{tab:ablation_data}

\end{table}

We also perform an ablation study of our training datasets.
We train our model with 1) all datasets listed in \tabref{data}, 2) all datasets except the two augmented datasets (CO3D augmented with Lumiere and static-view video data augmented with CAT3D), 3) synthetic 4D datasets only (Kubric and Objaverse).
For all three versions, we train the model for $60$k iterations, and evaluate its ability for separate camera and time control on the NSFF dataset~\cite{li2021neural}.

The quantitative results are presented in \tabref{ablation_data}.
While the numbers themselves do not show a large gap (as most of the evaluated pixels are static background), we observe more clear visual differences on in-the-wild data (see \figref{ablation_data}).
For the model trained only on synthetic 4D datasets, it already gives surprisingly good control over camera and time, but the generated scene motions are often unnatural and the generated novel views usually look worse.
This is likely a generalization issue.
For the model trained without the two augmented datasets, its main failure mode is ``fixed viewpoint, varying time'' --- in many cases, the camera still moves even when the model is instructed to only change the scene dynamics.
This is potentially caused by our imperfect filtering of the video data, where some videos of non-static viewpoints are treated as static-view videos.

\end{document}